\documentclass[english]{llncs}

\usepackage{textcomp}
\usepackage{times}
\usepackage{amssymb}
\usepackage{amsfonts}
\usepackage{latexsym}
\usepackage{url}
\usepackage{epsfig}
\usepackage{amssymb,amsmath}
\usepackage{amsfonts}
\usepackage{graphics}
\usepackage{graphicx}

\newcommand{\gl}[2]{{#1}^{{#2}}}

\newcommand{\la}{\langle}
\newcommand{\ra}{\rangle}
\newcommand{\st}{\,|\;}
\newcommand{\cA}{\mathcal{A}}

\newcommand{\cM}{\mathcal{M}}

\newcommand{\cSM}{\mathcal{ST}}

\newcommand{\cSP}{\mathcal{SP}}

\newcommand{\cL}{\mathcal{L}}

\newcommand{\vph}{\varphi}

\newcommand{\At}{\mathit{At}}
\newcommand{\Min}{\mathit{Min}}
\newcommand{\NP}{\mathit{NP}}

\newcommand{\tr}{\mathbf{t}}
\newcommand{\fa}{\mathbf{f}}
\newcommand{\un}{\mathbf{u}}
\newcommand{\rra}{\rightarrow}

\newcommand{\lla}{\leftarrow}

\newcommand{\n}{\mathit{not\;}}

\makeatother

\title{Revisiting Epistemic Specifications}

\author{
Miros{\l}aw Truszczy{\'n}ski
}
\institute{% 
Department of Computer Science\\
University of Kentucky\\
Lexington, KY~40506, USA\\
\email{mirek@cs.uky.edu}\\ 
}

\begin{document}
\maketitle

{\footnotesize
\hfill\parbox{2.5in}{
\emph{\hfill In honor of Michael Gelfond on his 65th birthday!}
}
}

\begin{abstract}
In 1991, Michael Gelfond introduced the language of epistemic 
specifications. The goal was to develop tools for modeling problems
that require some form of meta-reasoning, that is, reasoning over 
multiple possible worlds. Despite their relevance to knowledge
representation, epistemic specifications have received relatively 
little attention so far. In this paper, we revisit the formalism of 
epistemic specification. We offer a new definition of the formalism, 
propose several semantics (one of which, under syntactic restrictions
we assume, turns out to be equivalent to the original semantics by 
Gelfond), derive some complexity results and, finally, show the 
effectiveness of the formalism for modeling problems requiring 
meta-reasoning considered recently by Faber and Woltran. All these
results show that epistemic specifications deserve much more attention 
that has been afforded to them so far.
\end{abstract}

\section{Introduction}

Early 1990s were marked by several major developments in knowledge 
representation and nonmonotonic reasoning. One of the most important
among them was the introduction of \emph{disjunctive logic programs 
with classical negation} by Michael Gelfond and Vladimir Lifschitz 
\cite{gl90b}. The language of the formalism allowed for rules
\[
H_1 \vee \ldots \vee H_k \lla B_1,\ldots, B_m,\n B_{m+1},\ldots,\n B_n,
\]
where $H_i$ and $B_i$ are classical literals, that is, atoms and
classical or \emph{strong} negations ($\neg$) of atoms. In the
paper, we will write ``strong'' rather than ``classical'' negation,
as it reflects more accurately the role and the behavior of the 
operator.  The \emph{answer-set} semantics for programs consisting of 
such rules, introduced in the same paper, generalized the stable-model 
semantics of normal logic programs proposed a couple of years earlier 
also by Gelfond and Lifschitz \cite{gl88}. The proposed extensions of 
the language of normal logic programs were motivated by knowledge 
representation considerations. With two negation operators it was
straightforward to distinguish between $P$ being \emph{false by default}
(there is no justification for adopting $P$), and $P$ being 
\emph{strongly false} (there is evidence for $\neg P$). The former
would be written as $\n P$ while the latter as $\neg P$. And with the
disjunction in the head of rules one could model ``indefinite'' rules 
which, when applied, provide partial information only (one of the 
alternatives in the head holds, but no preference to any of them is 
given).

Soon after disjunctive logic programs with strong negation were 
introduced, Michael Gelfond proposed an additional important
extension, this time with a modal operator \cite{Gelfond91}. He called the
resulting formalism the language of \emph{epistemic specifications}. 
The motivation came again from knowledge representation. The goal 
was to provide means for the ``correct representation of incomplete 
information in the presence of multiple extensions'' \cite{Gelfond91}. 

Surprisingly, despite their evident relevance to the theory of
nonmonotonic
reasoning as well as to the practice of knowledge representation,
epistemic specifications have received relatively little attention 
so far. This
state of affairs may soon change. Recent work by Faber and Woltran
on \emph{meta-reasoning} 
with answer-set programming \cite{FaberW09} shows the need for languages,
in which one could express properties holding across all answer sets
of a program, something Michael Gelfond foresaw already two decades
ago.

Our goal in this paper is to revisit the formalism of epistemic 
specifications and show that they deserve a second look, in fact,
a place in the forefront of knowledge representation research. We will 
establish a general semantic framework for the formalism, and identify 
in it the precise location of Gelfond's epistemic specifications. We 
will derive several complexity results. We will also show that the 
original idea 
of Gelfond to use a modal operator to model ``what is known to a 
reasoner'' has a broader scope of applicability. In particular, we will 
show that it can also be used in combination with the classical logic. 

Complexity results presented in this paper provide an additional 
motivation to study epistemic specifications. Even though programs 
with strong negation often look ``more natural'' as they more directly 
align with the natural language description of knowledge 
specifications, the extension of the language of normal logic 
programs with the strong negation operator does not actually increase
the expressive power of the formalism. This point was made already by 
Gelfond and Lifschitz, who observed that there is a simple and concise 
way to compile the strong negation away. On the other hand, the 
extension allowing the disjunction operator in the heads of rules is 
an essential one. As the complexity results show \cite{mt88,eg95}, 
the class of problems that can be represented by means of disjunctive 
logic programs is strictly larger (assuming no collapse of the 
polynomial hierarchy) than the class of problems that can be modeled by 
normal logic programs. In the same vein, extension by the modal operator
along the lines proposed by Gelfond is essential, too. It does
lead to an additional jump in the complexity.

\section{Epistemic Specifications}

To motivate epistemic specifications, Gelfond discussed the following
example. A certain college has these rules to determine the eligibility
of a student for a scholarship:
\begin{enumerate}
\item Students with high GPA are eligible
\item Students from underrepresented groups and with fair GPA are eligible
\item Students with low GPA are not eligible
\item When these rules are insufficient to determine eligibility,
the student should be interviewed by the scholarship committee.
\end{enumerate}
Gelfond argued that there is no simple way to represent these rules
as a disjunctive logic program with strong negation. There is no problem
with the first three rules. They are modeled correctly by the following
three logic program rules (in the language with both the default and strong
negation operators):
\begin{enumerate}
\item $eligible(X) \lla \mathit{highGPA}(X)$
\item $eligible(X) \lla underrep(X), \mathit{fairGPA}(X)$
\item $\neg eligible(X) \lla \mathit{lowGPA}(X)$.
\end{enumerate}
The problem is with the fourth rule, as it has a clear meta-reasoning
flavor. It should apply when the possible worlds (answer sets)
determined by the first three rules do not fully specify the status of 
eligibility of a student $a$: neither \emph{all} of them contain 
$eligible(a)$ nor \emph{all} of them contain $\neg eligible(a)$. An 
obvious attempt at a formalization:
\begin{enumerate}
\item[4.] $interview(X) \lla \n eligible(X), \n \neg eligible(X)$ 
\end{enumerate}
fails. It is just another rule to be added to the program. Thus,
when the answer-set semantics is used, the rule is interpreted with
respect to individual answer sets and not with respect to collections
of answer-sets, as required for this application. For a concrete 
example, let us assume that all we know about a certain student named 
Mike is that Mike's GPA is fair or high. Clearly, we do not have 
enough information to determine Mike's eligibility and so we must 
interview Mike. But the program consisting of rules (1)-(4) and the 
statement 
\begin{enumerate}
\item[5.] $fairGPA(mike) \vee highGPA(mike)$
\end{enumerate}
about Mike's GPA, has two answer sets:
\begin{quote}
$\{highGPA(mike), eligible(mike)\}$\\
$\{fairGPA(mike), interview(mike)\}$. 
\end{quote}
Thus, the query $?interview(mike)$ has the answer ``unknown.'' To
address the problem, Gelfond proposed to extend the language with a
modal operator $K$ and, speaking informally, interpret premises 
$K\vph$ as ``$\vph$ is known to the program'' (the original phrase
used by Gelfond was ``known to the reasoner''), that is, true in all 
answer-sets. With this language extension, the 
fourth rule can be encoded as 
\begin{enumerate}
\item[4$'$.] $interview(X) \lla \n K\, eligible(X), \n K \neg eligible(X)$
\end{enumerate}
which, intuitively, stands for ``\emph{interview} if neither the 
eligibility nor the non-eligibility is known.'' 

The way in which Gelfond \cite{Gelfond91} proposed to formalize this 
intuition is strikingly elegant. We will now discuss it. We start with
the syntax of \emph{epistemic specifications}. As elsewhere in the 
paper, we restrict attention to the propositional case. We assume a 
fixed infinite countable set $\At$ of \emph{atoms} and the corresponding
language $\cL$ of propositional logic. A \emph{literal} is an atom, say
$A$, or its \emph{strong} negation $\neg A$. A \emph{simple modal atom}
is an expression $K\vph$, where $\vph\in \cL$, and a \emph{simple modal 
literal} is defined accordingly. An \emph{epistemic premise} is an 
expression (conjunction)
\[
E_1,\ldots, E_s,\n E_{s+1},\ldots, \n E_t,
\]
where every $E_i$, $1\leq i\leq t$, is a simple modal literal. 
An \emph{epistemic rule} is an expression of the form
\[
L_1 \vee \ldots\vee L_k \lla L_{k+1},\ldots, L_m,\n L_{m+1},\ldots, \n L_n, E,
\]
where every $L_i$, $1\leq i\leq k$, is a literal, and $E$ is an epistemic 
premise.
Collections of
epistemic rules are \emph{epistemic programs}. It is clear that (ground
versions of) rules (1)-(5) and (4$'$) are examples of epistemic rules, 
with rule (4$'$) being an example of an epistemic rule that actually 
takes advantage of the extended syntax. Rules such as
\begin{quote}
$a\vee \neg d \lla b, \n \neg c, \neg K(d\lor \neg c)$\\
$\neg a \lla \neg c, \n \neg K(\neg (a\land c)\rra b)$
\end{quote}
are also examples of epistemic rules. We note that the language of 
epistemic programs is only a fragment of the language of epistemic 
specifications by Gelfond. However, it is still expressive enough to 
cover all examples discussed by Gelfond and, more generally, a broad
range of practical applications, as natural-language formulations of 
domain knowledge typically assume a rule-based pattern.

We move on to the semantics, which is in terms of \emph{world 
views}. The definition of a world view consists of several steps. 
First, let $W$ be 
a consistent set of literals from $\cL$. We regard $W$ as a three-valued
interpretation of $\cL$ (we will also use the term \emph{three-valued 
possible world}), assigning to each atom one of the three logical values
$\tr$, $\fa$ and $\un$. The interpretation extends by recursion to all 
formulas in $\cL$, according to the following truth tables
\begin{figure}[h]
\begin{minipage}[t]{2.1cm}
\ 
\end{minipage}
\begin{minipage}[t]{1.5cm}
%\begin{table}[th]
\begin{center}
\begin{tabular}{|c|c|}
\hline
$\neg$ & \\
\hline
$\fa$  & $\tr$ \\
$\tr$ & $\fa$\\
$\un$ & $\un$\\
\hline
\end{tabular}
\end{center}
%\end{table}
\end{minipage}
\begin{minipage}[t]{2.0cm}
%\begin{table}[th]
\begin{center}
\begin{tabular}{|c||c|c|c|}
\hline
$\lor$ & $\tr$ & $\un$ & $\fa$ \\
\hline
\hline
$\tr$ & $\tr$ & $\tr$ & $\tr$ \\
$\un$ & $\tr$ & $\un$ & $\un$ \\
$\fa$ & $\tr$ & $\un$ & $\fa$ \\
\hline
\end{tabular}
\end{center}
\end{minipage}
\begin{minipage}[t]{2.0cm}
%\begin{table}[th]
\begin{center}
\begin{tabular}{|c||c|c|c|}
\hline
$\land$ & $\tr$ & $\un$ & $\fa$ \\
\hline
\hline
$\tr$ & $\tr$ & $\un$ & $\fa$ \\
$\un$ & $\un$ & $\un$ & $\fa$ \\
$\fa$ & $\fa$ & $\fa$ & $\fa$ \\
\hline
\end{tabular}
\end{center}
\end{minipage}
\begin{minipage}[t]{2.0cm}
%\begin{table}[th]
\begin{center}
\begin{tabular}{|c||c|c|c|}
\hline
$\rra$ & $\tr$ & $\un$ & $\fa$ \\
\hline
\hline
$\tr$ & $\tr$ & $\un$ & $\fa$ \\
$\un$ & $\tr$ & $\un$ & $\un$ \\
$\fa$ & $\tr$ & $\tr$ & $\tr$ \\
\hline
\end{tabular}
\end{center}
\end{minipage}\\ \\ \mbox{}
%\end{table}
\caption{Truth tables for the 3-valued logic of Kleene.}
\label{t.1}
\end{figure}

By a \emph{three-valued possible-world structure} we mean a non-empty
family of consistent sets of literals (three-valued possible worlds). 
Let $\cA$ be a three-valued possible-world structure and let $W$ be
a consistent set of literals. For every formula $\vph\in\cL$, we define 
\begin{enumerate}
\item $\la\cA,W\ra\models\vph$, if $v_W(\vph)=\tr$
\item $\la\cA,W\ra\models K\vph$, if for every $V\in\cA$, $v_V(\vph)=\tr$
\item $\la\cA,W\ra\models \neg K\vph$, if there is $V\in\cA$ such
that $v_V(\vph)=\fa$.
\end{enumerate}
Next, for every literal or simple modal literal $L$, we define
\begin{enumerate}
\item[4.] $\la\cA,W\ra \models\n L$ if $\la \cA,W\ra \not\models L$.
\end{enumerate}
We note that neither $\la\cA,W\ra\models K\vph$ nor $\la\cA,W\ra\models 
\neg K\vph$ depend on $W$. Thus, we will often write $\cA\models F$, when 
$F$ is a simple modal literal or its default negation.

In the next step, we introduce the notion of the \emph{G-reduct} of an 
epistemic program.

\begin{definition}
\label{def1}
Let $P$ be an epistemic program, $\cA$ a three-valued possible-world
structure and $W$ a consistent set of literals. The \emph{G-reduct}
of $P$ with respect to $\la\cA,W\ra$, in symbols $P^{\la\cA,W\ra}$, 
consists of the heads of all
rules $r\in P$ such that $\la\cA,W\ra\models \alpha$, for every 
conjunct $\alpha$ occurring in the body of $r$. 
\end{definition}

Let $H$ be a set of disjunctions of literals from $\cL$. A set $W$ of 
literals is \emph{closed} with respect to $H$ if $W$ is consistent
and contains at least one literal in common with every disjunction in 
$H$. 
%and if $W$ 
%consists of all literals if only it contains both $x$ and $\neg x$ 
%for some $x\in \At$. 
We denote by $\Min(H)$ the family of all minimal sets of literals that 
are closed with respect to $H$. %In other words, $\Min(H)$ consists of
%the set of all literals or of all minimal (according to the truth
%ordering) three-valued interpretations that satisfy every disjunction
%in $H$. 
With the notation $\Min(H)$ in hand, we are finally ready to
define the concept of a world view of an epistemic program $P$. 

\begin{definition}
\label{def2}
A three-valued possible-world structure $\cA$ is a \emph{world view} 
of an epistemic program $P$ if $\cA= \{W\st W\in \Min(P^{\la\cA,W\ra})\}$.
\end{definition}

\begin{remark}
The $G$-reduct of an epistemic program consists of disjunctions of literals.
Thus, the concept of a world view is well defined.
\end{remark}
\begin{remark}
We note that Gelfond considered also inconsistent sets of literals as 
minimal sets closed under disjunctions. However, the only such set 
he allowed consisted of \emph{all} literals. Consequently, the 
difference between the Gelfond's semantics and the one we described 
above is that some programs have a world view in the Gelfond's approach
that consists of a single set of all literals, while in our approach 
these programs do not have a world view. But in all other cases, the 
two semantics behave in the same way. 
\end{remark}

Let us consider the ground program, say $P$, corresponding to the 
scholarship eligibility example (rule (5), and rules (1)-(3) and 
(4$'$), grounded with respect to the Herbrand universe $\{mike\}$).
The only rule involving simple modal literals is
\begin{quote}
$interview(mike) \lla \n K\, eligible(mike), \n K \neg eligible(mike)$.
\end{quote}
Let $\cA$ be a world view of $P$. Being a three-valued possible-world 
structure, $\cA\not=\emptyset$. No
matter what $W$ we consider, no minimal set closed with respect to 
$P^{\la\cA,W\ra}$ contains $\mathit{lowGPA}(mike)$ and, consequently,
no minimal set closed with respect to $P^{\la\cA,W\ra}$ contains 
$\neg eligible(mike)$. It follows that $\cA\not\models K \neg 
eligible(mike)$. 

Let us assume that $\cA\models K\, eligible(mike)$. Then, no reduct
$P^{\la\cA,W\ra}$ contains $interview(mike)$. Let
$W=\{\mathit{fairGP}(mike)\}$. It follows that $P^{\la\cA,W\ra}$
consists only of $\mathit{fairGPA}(mike) \vee \mathit{highGPA}(mike)$.
Clearly, $W\in\Min(P^{\la\cA,W\ra})$ and, consequently, $W\in \cA$.
Thus, $\cA\not\models K\, eligible(mike)$, a contradiction.

It must be then that $\cA\models\n K\, eligible(mike)$ and 
$\cA\models \n K \neg eligible(mike)$. Let $W$ be an arbitrary 
consistent set of literals. Clearly, the reduct $P^{\la\cA,W\ra}$ 
contains $interview(mike)$ and
$\mathit{fairGPA}(mike) \vee \mathit{highGPA}(mike)$. If 
$\mathit{highGPA}(mike)\in W$, the reduct also contains $eligible(mike)$. 
Thus, $W\in\Min(P^{\la\cA,W\ra})$ if and only if 
\begin{quote}
$W=\{\mathit{fairGPA}(mike),interview(mike)\}$, or\\
$W=\{\mathit{highGPA}(mike),eligible(mike),interview(mike)\}$.
\end{quote}
It follows that if $\cA$ is a world view for $P$ then it consists of 
these two possible worlds. Conversely, it is easy to check that a 
possible-world structure consisting of these two possible worlds is a 
world view for $P$. Thus, $interview(mike)$ holds in $\cA$, and so 
our representation of the example as an epistemic program has the
desired behavior.

\section{Epistemic Specifications --- a Broader Perspective}

The discussion in the previous section demonstrates the usefulness of
formalisms such as that of epistemic specifications for knowledge
representation and reasoning. We will now present a simpler yet, in 
many respects, more general framework for epistemic specifications. 
The key to our approach is that we consider the semantics given by 
\emph{two-valued} interpretations (sets of atoms), and standard 
\emph{two-valued} possible-world structures (nonempty collections of 
two-valued interpretations). We also work within a rather standard
version of the language of modal propositional logic and so, in 
particular, we allow only for one negation operator. Later in the paper
we show that epistemic specifications by Gelfond can be encoded in a 
rather direct way in our formalism. Thus, the restrictions we impose 
are not essential even though, admittedly, not having two kinds of 
negation in the language in some cases may make the modeling task 
harder. 

We start by making precise the syntax of the language we will be 
using.  As we stated earlier, we assume a fixed infinite countable 
set of atoms $\At$. The language we consider is determined by the 
set $\At$, the modal operator $K$, and by the \emph{boolean connectives}
$\bot$ (0-place), and $\land$, $\lor$, and $\rra$ (binary). The BNF 
expression 
\begin{quote}
$\vph::= \bot\,|\,A\,|\,(\vph\land \vph)\,|\,(\vph \lor \vph)\,|\,
(\vph\rra\vph)\,|\ K\vph$,
\end{quote}
where $A\in \At$, provides a concise definition of a formula. The 
parentheses are used only to disambiguate the order of binary 
connectives. Whenever possible, we omit them. We define the unary
\emph{negation} connective $\neg$ and the 0-place connective $\top$
as abbreviations:
\begin{quote}
$\neg\vph::= \vph\rra\bot$\\
$\top::=\neg\bot$.
\end{quote}
We call formulas $K\vph$, where $\vph\in\cL_K$, \emph{modal atoms} 
(simple modal atoms that we considered earlier and will consider below
are special modal atoms with $K$-depth equal to 1). We denote this 
language by $\cL_K$ and refer to subsets of $\cL_K$ as \emph{epistemic 
theories}. We denote the modal-free fragment of $\cL_K$ by $\cL$.  

While we will eventually describe the semantics (in fact, several of 
them) for arbitrary epistemic theories, we start with an important
special case. Due to close analogies between the concepts we define 
below and the corresponding ones defined earlier in the context of the 
formalism of Gelfond, we ``reuse'' the terms used there. Specifically, 
by an \emph{epistemic premise} we mean a conjunction of simple modal 
literals. Similarly, by an \emph{epistemic rule} we understand an 
expression of the form
\begin{equation}
\label{eq10}
E\land L_1\land\ldots\land L_m \rra A_1\lor\ldots\lor A_n,
\end{equation}
where $E$ is an epistemic premise, $L_i$'s are literals (in $\cL$) and 
$A_i$'s are atoms. Finally, we call a collection of epistemic rules an 
\emph{epistemic program}. It will always be clear from the context, in 
which sense these terms are to be understood.

We stress that $\neg$ is not a primary connective in the language but a
derived one (it is a shorthand for some particular formulas involving
the rule symbol). Even though under some semantics we propose below this
negation operator has features of default negation, under some others it
does not. Thus, we selected for it the standard negation symbol $\neg$
rather than the ``loaded'' $\n$. 
  
A (two-valued) \emph{possible-world structure} is any nonempty family
$\cA$ of subsets of $\At$ (two-valued interpretations). In the remainder
of the paper, when we use terms ``interpretation'' and ``possible-world 
structure'' without any additional modifiers, we always mean a two-valued
interpretation and a two-valued possible-world structure.

Let $\cA$ be a possible-world structure and $\vph\in \cL$. We recall 
that $\cA\models K\vph$ precisely when $W\models \vph$, for every 
$W\in\cA$, and $\cA\models \neg K\vph$, otherwise. We will now define 
the \emph{epistemic reduct} of an epistemic program with respect to a
possible-world structure. 

\begin{definition}
Let $P\subseteq\cL_K$ be an epistemic program and let $\cA$ be a
possible-world structure. The \emph{epistemic reduct} of $P$ with 
respect to $\cA$, $\gl{P}{\cA}$ in symbols, is the theory obtained 
from $P$ as follows: eliminate every rule with an epistemic premise $E$
such that $\cA\not\models E$; drop the epistemic premise from every 
remaining rule. 
\end{definition}
It is clear that $\gl{P}{\cA}\subseteq\cL$, and that it consists of 
rules of the form 
\begin{equation}
\label{eq11}
L_1\land\ldots\land L_m \rra A_1\lor\ldots\lor A_n,
\end{equation}
where $L_i$'s are literals (in $\cL$) and $A_i$'s are atoms. 

Let $P$ be a collection of rules (\ref{eq11}). Then, $P$ is a 
propositional theory. Thus, it can be interpreted by the standard 
propositional logic semantics. However, $P$ can also be regarded as a 
disjunctive logic program (if we write rules from right to left rather 
than from left to right). Consequently, $P$ can also be interpreted by
the stable-model semantics \cite{gl88,gl90b} and the supported-model 
semantics \cite{abw87,bg93,bradix96jlp1,Inoue98-JLP}. (For normal logic
programs, the supported-model semantics was introduced by Apt et al. 
\cite{abw87}. The notion was extended to disjunctive logic programs by
Baral and Gelfond \cite{bg93}. We refer to papers by Brass and Dix
\cite{bradix96jlp1}, Definition 2.4, and Inoue and Sakama 
\cite{Inoue98-JLP}, Section 5, for more details).
We write $\cM(P)$, $\cSM(P)$ and 
$\cSP(P)$ for the sets of models, stable models and supported models 
of $P$, respectively. An important observation is that \emph{each} of 
these semantics gives rise to the corresponding notion of an epistemic
extension.

\begin{definition}
\label{def11}
Let $P\subseteq \cL_K$ be an epistemic program. A possible-world 
structure $\cA$ is an \emph{epistemic model} (respectively, an
\emph{epistemic stable model}, or an \emph{epistemic supported model})
of $P$, if $\cA = \cM(\gl{P}{\cA})$ (respectively, $\cA = 
\cSM(\gl{P}{\cA})$ or $\cA = \cSP(\gl{P}{\cA})$).
\end{definition}

It is clear that Definition \ref{def11} can easily be adjusted also to
other semantics of propositional theories and programs. We briefly 
mention two such semantics in the last section of the paper.

We will now show that epistemic programs with the semantics of 
epistemic stable models can provide an adequate representation to
the scholarship eligibility example for Mike. The available information
can be represented by the following program $P(mike) \subseteq\cL_K$:

\begin{enumerate}
\item $eligible(mike)\land neligible(mike) \rra \bot$
\item $fairGPA(mike) \vee highGPA(mike)$
\item $\mathit{highGPA}(mike) \rra eligible(mike)$
\item $underrep(mike) \land \mathit{fairGPA}(mike)\rra eligible(mike)$
\item $\mathit{lowGPA}(mike) \rra neligible(mike)$
\item $\neg K\, eligible(mike), \neg K\, neligible(mike)\rra interview(mike)$.
\end{enumerate}
We use the predicate \emph{neligible} to model the strong negation 
of the predicate $eligible$ that appears in the representation in terms 
of epistemic programs by Gelfond (thus, in particular, the presence of 
the first clause, which precludes the facts $eligible(mike)$ and 
$neligible(mike)$
to be true together). This extension of the language and an extra
rule in the representation is the price we pay
for eliminating one negation operator. 

Let $\cA$ consist of the interpretations 
\begin{quote}
$W_1=\{\mathit{fairGPA}(mike),interview(mike)\}$\\
$W_2=\{\mathit{highGPA}(mike),eligible(mike),interview(mike)\}$.
\end{quote}
Then the reduct $\gl{[P(mike)]}{\cA}$ consists of rules (1)-(5),
which are unaffected by the reduct operation, and of the fact 
$interview(mike)$, resulting from rule (6) when the reduct operation
is performed (as in logic programming, when a rule has the empty 
antecedent, we drop the implication symbol from the 
notation). One can check that 
$\cA=\{W_1,W_2\}=\cSM(\gl{[P(mike)]}{\cA})$. Thus, $\cA$ is
an epistemic stable model of $P$ (in fact, the only one). Clearly, 
$interview(mike)$ holds in the model (as we would expect it to), as it 
holds in each of its possible-worlds. We note that in this particular 
case, the semantics of epistemic supported models yields exactly the 
same solution.

\section{Complexity}

We will now study the complexity of reasoning with epistemic (stable, 
supported) models. We provide details for the case of epistemic stable
models, and only present the results for the other two semantics, as the 
techniques to prove them are very similar to those we develop for the
case of epistemic stable models.

First, we note that epistemic stable models of an epistemic program $P$
can be represented by partitions of the set of all modal atoms of $P$. 
This is important as \emph{a priori} the size of possible-world 
structures one needs to consider as candidates for epistemic stable
models may be exponential in the size of a program. Thus, to obtain 
good complexity bounds alternative polynomial-size representations 
of epistemic stable models are needed.

Let $P\subseteq\cL_K$ be an epistemic program and $(\Phi,\Psi)$ be the 
set of modal atoms of $P$ (all these modal atoms are, in fact, simple).
We write $P_{|\Phi,\Psi}$ for the program
obtained from $P$ by eliminating every rule whose epistemic premise
contains a conjunct $K\psi$, where $K\psi\in\Psi$, or a conjunct
$\neg K\vph$, where $K\vph\in\Phi$ (these rules are ```blocked'' by
$(\Phi,\Psi)$), and by eliminating the epistemic premise from every
other rule of $P$.

\begin{proposition}
\label{prop:char}
Let $P\subseteq\cL_K$ be an epistemic program. If a possible-world
structure $\cA$ is an epistemic stable model of $P$, then there
is a partition $(\Phi,\Psi)$ of the set of modal atoms of $P$ such
that
\begin{enumerate}
\item $\cSM(P_{|\Phi,\Psi})\not=\emptyset$
\item For every $K\vph\in\Phi$, $\vph$ holds in every stable model
of $P_{|\Phi,\Psi}$ 
\item For every $K\psi\in\Psi$, $\psi$ does not hold in at least one
stable model of $P_{|\Phi,\Psi}$. 
\end{enumerate}
Conversely, if there are such partitions, $P$ has epistemic stable 
models.
\end{proposition}  

It follows that epistemic stable models can be represented by partitions
$(\Phi,\Psi)$ satisfying conditions (1)-(3) from the proposition above. 

We observe that deciding whether a partition $(\Phi,\Psi)$ satisfies 
conditions (1)-(3) from Proposition \ref{prop:char}, can be accomplished
by polynomially many calls to an $\Sigma_2^P$-oracle and, if we restrict
attention to non-disjunctive epistemic programs, by polynomially many 
calls to an $\NP$-oracle. 

\begin{remark}
\label{rem1}
If we adjust Proposition \ref{prop:char} by replacing the term ``stable''
with the term ``supported,'' and replacing $\cSM()$ with $\cSP()$, we
obtain a characterization of epistemic supported models. Similarly, 
omitting the term ``stable,'' and replacing $\cSM()$ with $\cM()$ 
yields a characterization of epistemic models. In each case, one can 
decide whether a partition $(\Phi,\Psi)$ satisfies conditions (1)-(3)
by polynomially many calls to an $\NP$-oracle (this claim is evident 
for the case of epistemic models; for the case of epistemic supported
models, it follows from the fact that supported models semantics does
not get harder when we allow disjunctions in the heads or rules). 
\end{remark}

\begin{theorem}
The problem to decide whether a non-disjunctive epistemic program has 
an epistemic stable model is $\Sigma_2^P$-complete.  
\end{theorem}
Proof: Our comments above imply that the problem is in the class 
$\Sigma_2^P$. Let $F=\exists Y \forall Z \Theta$, where $\Theta$
is a DNF formula. The problem to decide whether $F$ is true is 
$\Sigma_2^P$-complete. We will reduce it to the problem in question 
and, consequently, demonstrate its $\Sigma_2^P$-hardness. To this
end, we construct an epistemic program $Q\subseteq\cL_K$ by including 
into $Q$ the following clauses (atoms $w$, $y'$, $y\in Y$, and $z'$, 
$z\in Z$ are fresh):
\begin{enumerate}
\item $Ky\rra y\;$; and $Ky' \rra y'$, for every $y\in Y$
\item $y\land y'\rra\;$; and  $\neg y\land \neg y'\rra\;$, for every $y\in Y$
\item $\neg z'\rra z\;$; and $\neg z\rra z'$, for $z\in Z$
\item $\sigma(u_1)\land  \ldots\land  \sigma(u_k)\rra w\;$, where 
$u_1\wedge\ldots\wedge u_k$ is a disjunct of $\Theta$, and 
$\sigma(\neg a)=a'$ and $\sigma(a)=a$, for every $a\in Y\cup Z$
\item $\neg Kw\rra\;$.
\end{enumerate}

Let us assume that $\cA$ is an epistemic stable model of $Q$. In particular,
$\cA\not=\emptyset$. It must 
be that $\cA\models Kw$ (otherwise, $\gl{Q}{\cA}$ has no stable models, 
that is, $\cA=\emptyset$). Let us define $A= \{y\in Y\st \cA \models 
Ky\}$, and $B=\{y\in Y\st\cA\models Ky'\}$. It follows that $\gl{Q}{\cA}$
consists of the following rules:
\begin{enumerate}
\item $y$, for $y\in A$, and $y'$, for $y\in B$
\item $y\land y'\rra\;$; and  $\neg y\land \neg y'\rra \;$, for every $y\in Y$
\item $\neg z'\rra z\;$; and $\neg z\rra z'$, for $z\in Z$
\item $\sigma(u_1)\land  \ldots\land  \sigma(u_k)\rra w\;$, where
$u_1\wedge\ldots\wedge u_k$ is a disjunct of $\Theta$, and
$\sigma(\neg a)=a'$ and $\sigma(a)=a$, for every $a\in Y\cup Z$.
\end{enumerate}
Since $\cA=\cSM(\gl{Q}{\cA})$ and $\cA\not=\emptyset$, $B=Y\setminus A$
(due to clauses of type (2)).
It is clear that the program $\gl{Q}{\cA}$ has stable models
and that they are of the form $A\cup \{y'\st y\in Y\setminus A\} \cup
D \cup \{z'\st z\in Z\setminus D\}$, if that set does not imply $w$ through
a rule of type (4), or $A\cup \{y'\st y\in Y\setminus A\} \cup
D \cup \{z'\st z\in Z\setminus D\}\cup \{w\}$, otherwise, where $D$ is any
subset of $Z$. As $\cA\models Kw$,
there are no stable models of the first type. Thus, the family of stable 
models
of $\gl{Q}{\cA}$ consists of all sets $A\cup \{y'\st y\in Y\setminus 
A\} \cup
D \cup \{z'\st z\in Z\setminus D\}\cup \{w\}$, where $D$ is an arbitrary
subset of $Z$. It follows that for every $D\subseteq Z$, the set $A\cup 
\{y'\st y\in Y\setminus A\} \cup D \cup \{z'\st z\in Z\setminus D\}$
satisfies the body of at least one rule of type (4). By the construction, 
for every $D\subseteq Z$, the valuation of $Y\cup Z$ determined by $A$ and 
$D$ satisfies the corresponding disjunct in $\Theta$ and so, also $\Theta$.
In other words, $\exists Y\forall Z \Theta$ is true.

Conversely, let $\exists Y\forall Z\Theta$ be true. Let $A$ be a subset
of $Y$ such that $\Theta_{|Y/A}$ holds for every truth assignment of $Z$ 
(by $\Theta_{|Y/A}$, we mean the formula obtained by simplifying the 
formula $Q$ with respect to the truth assignment of $Y$ determined by 
$A$). Let 
$\cA$ consist of all sets of the form $A\cup \{y'\st y\in Y\setminus 
A\} \cup D \cup \{z'\st z\in Z\setminus D\}\cup \{w\}$, where $D\subseteq
Z$. It follows that $\gl{Q}{\cA}$ consists of clauses (1)-(4) above,
with $B=Y\setminus A$.
%\begin{enumerate}
%\item $y$, $y\in A$ and $y'$, $y\in B$
%\item $\lla y,y'$; and  $\lla \n y,\n y'$, for every $y\in Y$
%\item $z \lla \n z'$; and $z'\lla \n z$, for $z\in Z$
%\item $w \lla \sigma(u_1), \ldots, \sigma(u_k)$, where $u_1\wedge\ldots\wedge
%u_k$ is a disjunct of $\Theta$, and $\sigma(\neg a)=a'$ and $\sigma(a)=a$,
%for every $a\in Y\cup Z$.
%\end{enumerate}
Since $\forall Z \Theta_{|A/Y}$ holds, it follows that $\cA$ is precisely 
the set of stable models of $\gl{Q}{\cA}$. Thus, $\cA$ is an epistemic
stable model of $Q$. 
\hfill$\Box$

In the general case, the complexity goes one level up.

\begin{theorem}
\label{thm:2}
The problem to decide whether an epistemic program $P\subseteq\cL_K$
has an epistemic stable model is $\Sigma_3^P$-complete.
\end{theorem}
Proof: The membership follows from the earlier remarks. To prove the 
hardness part, we consider a QBF formula $F=
\exists X \forall Y \exists Z \Theta$, where $\Theta$ is a 3-CNF 
formula. For each atom $x\in X$ ($y\in Y$ and $z\in Z$, respectively), we 
introduce a fresh atom $x'$ ($y'$ and $z'$, respectively). Finally, 
we introduce three additional fresh atoms, $w$, $f$ and $g$.

We now construct a disjunctive epistemic program $Q$ by including into
it the following clauses:
\begin{enumerate}
\item $Kx\rra x$; and $Kx' \rra x'$, for every $x\in X$
\item $x\land x'\rra$; and  $\neg x\land \neg x'\rra$, for every $x\in X$
\item $\neg g\rra f$; and $\neg f\rra g$
\item $f\rra y \vee y'$; and $f\rra z \vee z'$, for every $y\in Y$ 
and $z\in Z$
\item $f\land w\rra z$; and $f\land w\rra z'$, for every $z\in Z$
\item $f\land \sigma(u_1)\land \sigma(u_2)\land \sigma(u_3)\rra w$, 
for every clause $C= u_1\vee u_2 \vee u_3$ of $\Theta$, where 
$\sigma(a)=a'$ and 
$\sigma(\neg a)= a$, for every $a\in X\cup Y \cup Z$  
\item $f\land \neg w \rra w$
\item $\neg K\neg w \rra$
\end{enumerate}
Let us assume that $\exists X \forall Y\exists Z \Theta$ is true. Let
$A\subseteq
X$ describe the truth assignment on $X$ so that $\forall Y
\exists Z \Phi_{X/A}$ holds (we define $\Phi_{X/A}$ in the proof
of the previous result). We will show that $Q$ has an epistemic 
stable model
$\cA=\{A\cup \{a'\st a\in X\setminus A\}\cup\{g\}\}$. Clearly, $Kx$, $x\in A$,
and $Kx'$, $x\in X\setminus A$, are true in $\cA$. Also, $K\neg w$ is 
true in $\cA$. All other modal atoms in $Q$ are false in $\cA$.
Thus, $\gl{Q}{\cA}$ consists of rules $x$, for $x\in A$, $x'$, for $x\in
X\setminus A$ and of rules (2)-(7) above. Let $M$ be a stable model of
$\gl{Q}{\cA}$ containing $f$. It follows that $w\in M$ and so, $Z\cup Z'
\subseteq M$. Moreover, the Gelfond-Lifschitz reduct of $\gl{Q}{\cA}$
with respect to $M$ consists of rules $x$, for $x\in A$, $x'$, for $x\in
X\setminus A$, all $\neg$-free constraints of type (2), rule $f$, and 
rules (4)-(6) above, and $M$ is a minimal model of this program.

Let $B=Y\cap M$. By the minimality of $M$, $M=A\cup \{x'\st x\in X
\setminus A\}\cup B\cup\{y'\st y\in Y\setminus B\} \cup Z\cup Z'\cup 
\{f,w\}$. Since $\forall Y \exists Z \Phi_{X/A}$ holds, $\exists Z 
\Phi_{X/A,Y/B}$ holds, too. Thus, let $D\subseteq Z$ be a subset of $Z$
such that $\Phi_{X/A,Y/B,Z/D}$ is true. It follows that $M'=A\cup \{x'
\st x\in X \setminus A\}\cup B\cup\{y'\st y\in Y\setminus B\} \cup 
D\cup\{z'\st z\in Z\setminus D\} \cup \{f\}$ is also a model of the
Gelfond-Lifschitz reduct of $\gl{Q}{\cA}$ with respect to $M$, 
contradicting the minimality of $M$.

Thus, if $M$ is an answer set of $\gl{Q}{\cA}$, it must contain $g$. 
Consequently, it does not contain $f$ and so no rules of type (4)-(7) 
contribute to it. It follows that $M=A\cup \{a'\st a\in 
X\setminus A\} \cup\{g\}$ and, as it indeed is an answer set of 
$\gl{Q}{\cA}$, $\cA=\cSM(\gl{Q}{\cA})$. Thus, $\cA$ is a epistemic
stable model, as claimed.

Conversely, let as assume that $Q$ has an epistemic stable model, say,
$\cA$. It must be that $\cA\models K\neg w$ (otherwise, $\gl{Q}{\cA}$ 
contains a contradiction and has no stable models). Let us define 
$A=\{x\in X\st \cA\models Kx\}$ and $B=\{x\in X\st \cA\models Kx'\}$. 
It follows that $\gl{Q}{\cA}$ consists of the clauses: 
\begin{enumerate}
\item $x$, for $x\in A$ and $x'$, for $x\in B$
\item $x\land x'\rra $; and  $\neg x\land \neg x'\rra$, for every $x\in X$
\item $\neg g\rra f$; and $\neg f\rra g$
\item $f \rra y \vee y'$; and $f\rra z \vee z'$, for every $y\in Y$
and $z\in Z$
\item $f\land w \rra z$; and $f\land w \rra z'$, for every $z\in Z$
\item $f\land \sigma(u_1)\land\sigma(u_2)\land\sigma(u_3)\rra w$, for 
every clause $C= u_1\vee u_2 \vee u_3$ of $\Phi$, where $\sigma(a)=a'$
and $\sigma(\neg a)= a$, for every $a\in X\cup Y \cup Z$.
\item $f, \neg w \rra w$
\end{enumerate}
We have that $\cA$ is precisely the set of stable models of this 
program. Since $\cA\not=\emptyset$, $B=X\setminus A$. If $M$ is 
a stable model of $\gl{Q}{\cA}$ and contains $f$, then it contains 
$w$. But then, as $M\in\cA$, $\cA\not\models K\neg w$, a 
contradiction. It follows that there is no stable model containing
$f$. That is, the program consisting of the following rules has no
stable model:
\begin{enumerate}
\item $x$, for $x\in A$ and $x'$, for $x\in X\setminus A$
\item $y \vee y'$; and $z \vee z'$, for every $y\in Y$ and $z\in Z$
\item $w \rra z$; and $w \rra z'$, for every $z\in Z$
\item $\sigma(u_1)\land\sigma(u_2)\land\sigma(u_3)\rra w$, for
every clause $C= u_1\vee u_2 \vee u_3$ of $\Theta$, where $\sigma(a)=a'$
and $\sigma(\neg a)= a$, for every $a\in X\cup Y \cup Z$.
\item $\neg w \rra w$
\end{enumerate}
But then, the formula $\forall Y\exists Z \Theta_{|X/A}$ is true and,
consequently, the formula $\exists X \forall Y\exists Z \Theta$ is
true, too.  \hfill$\Box$

For the other two epistemic semantics, Remark 1 implies that the problem
of the existence of an epistemic model (epistemic supported model) is
in the class $\Sigma_2^P$. The $\Sigma_2^P$-hardness of the problem
can be proved by similar techniques as those we used for the case of
epistemic stable models. Thus, we have the following result.

\begin{theorem}
\label{thm:3}
The problem to decide whether an epistemic program $P\subseteq\cL_K$
has an epistemic model (epistemic supported model, respectively) is
$\Sigma_2^P$-complete.
\end{theorem}

\section{Modeling with Epistemic Programs}

We will now present several problems which illustrate the 
advantages offered by the language of epistemic programs we developed
in the previous two sections. Whenever we use predicate programs, we 
understand that their semantics is that of the corresponding ground 
programs.

First, we consider two graph problems related to the existence of 
Hamiltonian cycles. Let $G$ be a directed graph. An edge in $G$ is 
\emph{critical} if it belongs to every hamiltonian cycle in $G$. 
The following problems are of interest:
\begin{enumerate}
\item Given a directed graph $G$, find the set of all critical 
edges of $G$
\item Given a directed graph $G$, and integers $p$ and $k$, find a set
$R$ of no more than $p$ new edges such that $G\cup R$ has no more than
$k$ critical edges.
\end{enumerate}

Let $HC(vtx,edge)$ be any standard ASP encoding of the Hamiltonian 
cycle problem, in which predicates $vtx$ and $edge$ represent $G$, 
and a predicate $hc$ represents edges of a candidate hamiltonian 
cycle. We assume the rules of $HC(vtx,edge)$ are written from left
to right so that they can be regarded as elements of $\cL$. Then, 
simply adding to $HC(vtx,edge)$ the rule:
\begin{quote}
$K hc(X,Y) \rra critical(X,Y)$
\end{quote}
yields a correct representation of the first problem. We write
$HC_{cr}(vtx,edge)$ to denote this program.
Also, for a directed graph $G=(V,E)$,
we define 
\begin{quote}
$D=\{vtx(v)\st v\in V\} \cup \{edge(v,w)\st (v,w)\in E\}$.
\end{quote}
We have the following result.

\begin{theorem}
Let $G=(V,E)$ be a directed graph. If $HC_{cr}(vtx,edge) \cup D$ has 
no epistemic stable models, then every edge in $G$ is critical 
(trivially). Otherwise, the epistemic program
$HC_{cr}(vtx,edge)\cup D$ has a unique 
epistemic stable model $\cA$ and the set $\{(v,w)\st \cA\models 
critical(u,v) \}$ is the set of critical edges in $G$.
\end{theorem}
Proof (Sketch): Let $H$ be the grounding of $HC_{cr}(vtx,edge) \cup D$.
If $H$ has no epistemic stable models, it follows that the ``non-epistemic''
part $H'$ of $H$ has no stable models (as no atom of the form 
$critical(x,y)$ appears in it). As $H'$ encodes the existence of a hamiltonian
cycle in $G$, it follows that $G$ has no Hamiltonian cycles. Thus, trivially,
every edge of $G$ belongs to every Hamiltonian cycle of $G$ and so, every
edge of $G$ is critical.

Thus, let us assume that $\cA$ is an epistemic stable model of $H$. 
Also, let $S$ be the set of all stable models of $H'$ (they correspond 
to Hamiltonian cycles of $G$; each model contains, in particular, atoms 
of the form $hc(x,y)$, where $(x,y)$ ranges over the edges of the 
corresponding Hamiltonian cycle). The reduct $\gl{H}{\cA}$ consists of 
$H'$ (non-epistemic part of $H$ is unaffected by the reduct operation) 
and of $C'$, a set of some facts of the form $critical(x,y)$. Thus, the 
stable models of the reduct are of the form $M\cup C'$, where $M\in S$.
That is, $\cA=\{M\cup C'\st M\in S\}$. Let us denote by $C$ the set
of the atoms $critical(x,y)$, where $(x,y)$ belongs to every hamiltonian 
cycle of $G$ (is critical). One can compute now that $\gl{H}{\cA} = 
H'\cup C$. Since $\cA=\cSM(\gl{H}{\cA})$, $\cA= \{M\cup C\st M\in S\}$. 
Thus, $HC_{cr}(vtx,edge)\cup D$ has a unique epistemic stable model, as 
claimed. It also follows that the set $\{(v,w)\st \cA\models 
critical(u,v) \}$ is the set of critical edges in $G$.
\hfill$\Box$

To represent the second problem, we proceed as follows. First, we
``select'' new edges to be added to the graph and impose constraints
that guarantee that all new edges are indeed new, and that no more than $p$
new edges are selected (we use here \emph{lparse} syntax for brevity; the
constraint can be encoded strictly in the language $\cL_K$).
\begin{quote}
$vtx(X)\land vtx(Y) \rra newEdge(X,Y)$\\
$newEdge(X,Y)\land edge(X,Y)\rra\bot$\\
$(p+1) \{newEdge(X,Y): vtx(X), vtx(Y)\} \rra \bot$.
\end{quote}
%or
%\begin{quote}
%$\{newEdge(X,Y): vtx(X), vtx(Y), not edge(X,Y)\} p.$
%\end{quote}
Next, we define the set of edges of the extended graph, using a predicate
$edgeEG$:
\begin{quote}
$edge(X,Y)\rra edgeEG(X,Y)$\\
$newEdge(X,Y)\rra edgeEG(X,Y)$
\end{quote}
Finally, we define critical edges and impose a constraint on their 
number (again, exploiting the \emph{lparse} syntax for brevity sake):
\begin{quote}
$edgeEG(X,Y)\land K hc(X,Y) \rra critical(X,Y)$\\
$(k+1)\{critical(X,Y): edgeEG(X,Y)\}\rra \bot$.
\end{quote}
We define $Q$ to consist of all these rules together with all the rules
of the program $HC(vtx,edgeEG)$. We now have the following theorem. The
proof is similar to that above and so we omit it.
 
\begin{theorem}
Let $G$ be a directed graph. There is an extension of $G$ with no more
than $p$ new edges so that the resulting graph has no more than $k$ 
critical edges if and only if the program $Q\cup D$ has an epistemic 
stable model.
\end{theorem}
 
For another example we consider the unique model problem: given a CNF 
formula $F$, the goal is to decide whether $F$ has a unique minimal 
model. The unique model problem was also considered by Faber and Woltran
\cite{FaberW09}. We will show two encodings of the problem by means of
epistemic programs. The first one uses the semantics of epistemic 
models and is especially direct. The other one uses the semantics of
epistemic stable models. 

Let $F$ be a propositional theory consisting of constraints $L_1\land
\ldots\land L_k\rra \bot$, where $L_i$'s are literals. Any propositional
theory can be rewritten into an equivalent theory of such form. We 
denote by $F^K$ the formula obtained from $F$ by replacing every atom 
$x$ with the modal atom $Kx$. 

\begin{theorem}
For every theory $F\subseteq\cL$ consisting of constraints, $F$ has a 
least model if and only if the epistemic program $F\cup F^K$ has an
epistemic model.
\end{theorem}
Proof: Let us assume that $F$ has a least model. We define $\cA$ to
consist of all models of $F$, and we denote the least model of $F$ by 
$M$. We will show that $\cA$ is an epistemic model of $F\cup F^K$. 
Clearly, for every $x\in M$, $\cA\models Kx$. Similarly, for every 
$x\not\in M$, $\cA\models \neg Kx$. Thus, $\gl{[F^K]}{\cA}=\emptyset$.
Consequently, $\gl{[F\cup F^K]}{\cA} = F$ and so, $\cA$ is precisely 
the set of all models of $\gl{[F\cup F^K]}{\cA}$. Thus, $\cA$ is an 
epistemic model.  

Conversely, let $\cA$ be an epistemic model of $F\cup F^K$. It follows
that $\gl{[F^K]}{\cA}=\emptyset$ (otherwise, $\gl{[F\cup F^K]}{\cA}$ 
contains $\bot$ and $\cA$ would have to be empty, contradicting the definition
of an epistemic model). Thus, $\gl{[F\cup F^K]}{\cA}=F$ and consequently, 
$\cA$ is the set of all
models of $F$. Let $M=\{x\in\At\st \cA\models Kx\}$ and let
\begin{equation}
\label{eq17}
a_1\land\ldots\land a_m \land\neg b_1\land\ldots\land\neg b_n \rra \bot
\end{equation}
be a rule in $F$. Then, 
\[
K a_1\land\ldots\land K a_m \land\neg K b_1\land\ldots\land\neg K b_n
\rra \bot
\]
is a rule in $F^K$. As $\gl{[F^K]}{\cA}=\emptyset$,
\[
\cA\not\models K a_1\land\ldots\land K a_m \land\neg K b_1\land\ldots\land
\neg K b_n.
\]
Thus, for some $i$, $1\leq i\leq m$, $\cA\not\models K a_i$, or for 
some $j$, $1\leq j\leq n$, $\cA\models K b_j$. In the first case, $a_i
\notin M$, in the latter, $b_j\in M$. In either case, $M$ is a model
of rule (\ref{eq17}). It follows that $M$ is a model of $F$. Let $M'$
be a model of $F$. Then $M'\in\cA$ and, by the definition of $M$, $M
\subseteq M'$. Thus, $M$ is a least model of $F$.
\hfill$\Box$
 
Next, we will encode the same problem as an epistemic program under the
epistemic stable model semantics. The idea is quite similar. We
only need to add rules to generate all candidate models.

\begin{theorem}
For every theory $F\subseteq\cL$ consisting of constraints, $F$ has a
least model if and only if the epistemic program 
\[
F\cup F^K \cup \{\neg x\rra x'\st x\in \At\}\cup \{\neg x'\rra x\st 
x\in \At\}
\]
has an epistemic stable model.
\end{theorem}

We note that an even simpler encoding can be obtained if we use 
\emph{lparse} choice rules. In this case, we can replace 
$\{\neg x\rra x'\st x\in \At\}\cup \{\neg x'\rra x\st x\in \At\}$ with
$\{\{x\}\st x\in\At\}$.

\section{Connection to Gelfond's Epistemic Programs}

We will now return to the original formalism of epistemic specifications
proposed by Gelfond \cite{Gelfond91} (under the restriction to epistemic
programs we discussed here). We will show that it can be expressed in a
rather direct way in terms of our epistemic programs in the two-valued
setting and under the epistemic supported-model semantics.
 
The reduction we are about to describe is similar to the well-known
one used to eliminate the ``strong'' negation from disjunctive logic
programs with strong negation. In particular, it requires an extension 
to the language $\cL$. Specifically, for every atom $x\in \At$ we 
introduce a fresh atom $x'$ and we denote the extended language by 
$\cL'$. The intended role of $x'$ is to represent in $\cL'$ the literal 
$\neg x$ from $\cL$. Building on this idea, we assign to each set $W$ of
literals in $\cL$ the set
\[
W'=(W\cap \At) \cup \{x'\st \neg x\in W\}.
\]
In this way, sets of literals from $\cL$ (in particular, three-valued
interpretations of $\cL$) are represented as sets of atoms from $\cL'$
(two-valued interpretations of $\cL'$).

We now note that the truth and falsity of a formula form $\cL$ under
a three-valued interpretation can be expressed as the truth and
falsity of certain formulas from $\cL'$ in the two-valued setting.
The following result is well known.

\begin{proposition}
\label{prop:2}
For every formula $\vph\in\cL$ there are formulas
$\vph^-, \vph^+\in\cL'$ such that for every set of literals $W$
(in $\cL$)
\begin{enumerate}
\item $v_W(\vph)=\tr$ if and only if $u_{W'}(\vph^+)=\tr$
\item $v_W(\vph)=\fa$ if and only if $u_{W'}(\vph^-)=\fa$
\end{enumerate}
Moreover, the formulas $\vph^-$ and $\vph^+$ can be constructed in 
polynomial time with respect to the size of $\vph$.
\end{proposition}
Proof: This a folklore result. We provide a sketch of a proof for the
completeness sake. We define $\vph^+$ and $\vph^-$ by recursively as
follows:
\begin{enumerate}
\item $x^+ = x$ and $x^- = \neg x'$, if $x\in\At$
\item $(\neg \vph)^+ = \neg\vph^-$ and $(\neg \vph)^- = \neg\vph^+$
\item $(\vph\lor \psi)^+=\vph^+\lor \psi^+$ and 
$(\vph\lor \psi)^-=\vph^-\lor \psi^-$; the case of the conjunction is dealt
with analogously
\item $(\vph\rra\psi)^+ = \vph^-\rra\psi^+$ and 
$(\vph\rra\psi)^- = \vph^+\rra\psi^-$.
\end{enumerate}
One can check that formulas $\vph^+$ and $\vph^-$ defined in this way satisfy
the assertion. \hfill$\Box$

\smallskip
We will now define the transformation $\sigma$ that allows us to eliminate
strong negation. First, for a literal $L\in \cL$, we now define
\[
\sigma(L) = \left\{ \begin{array}{ll}
                    x     & \mbox{if $L=x$}\\
                    x'      & \mbox{if $L=\neg x$}
                    \end{array}
            \right.
\]
Furthermore, if $E$ is a simple modal literal or its default negation, 
we define
\[
\sigma(E) = \left\{ \begin{array}{ll}
                    K\vph^+     & \mbox{if $E=K\vph$}\\
                    \neg K\vph^{-}      & \mbox{if $E=\neg K\vph$}\\
                    \neg K\vph^{+}      & \mbox{if $E=\n K\vph$}\\
                    K\vph^{-}      & \mbox{if $E=\n \neg K\vph$}
                    \end{array}
            \right.
\]
and for an epistemic premise $E = E_1,\ldots, E_t$
(where each $E_i$ is a simple modal literal or its default negation) we set
\[
\sigma(E) = \sigma(E_1)\land\ldots\land \sigma(E_t).
\]
Next, if $r$ is an epistemic rule
\[
L_1 \vee \ldots\vee L_k \lla F_1,\ldots, F_m,\n F_{m+1},\ldots, \n F_n, E
\]
we define
\[
\sigma(r) = \sigma(E)\land \sigma(F_1)\land\ldots\land \sigma(F_m)\land \neg
\sigma(F_{m+1})\land\ldots\land \neg \sigma(F_n) \rra 
\sigma(L_1) \vee \ldots\vee \sigma(L_k).
\]
Finally, for an epistemic program $P$, we set 
\[
\sigma(P)=\{\sigma(r)\st r\in P\}) \cup\{x\land x' \rra \bot\}.
\]
We note that $\sigma(P)$ is indeed an epistemic program
in the language $\cL_K$ (according to our definition of epistemic 
programs). The role of the rules $x\land x' \rra \bot$
is to ensure that sets forming epistemic (stable, supported) models 
of $\sigma(P)$ correspond to consistent sets of literals (the only type 
of set of literals allowed in world views).

Given a three-valued possible structure $\cA$, we define $\cA'=\{W'\st
W\in \cA\}$, and we regard $\cA'$ as a two-valued possible-world
structure. We now have the following theorem.

\begin{theorem}
Let $P$ be an epistemic program according to Gelfond. Then a three-valued
possible-world structure $\cA$ is a world view of $P$ if and only if 
a two-valued possible-world structure $\cA'$ is an epistemic supported
model of $\sigma(P)$.
\end{theorem}
Proof (Sketch): Let $P$ be an epistemic program according to Gelfond, $\cA$
a possible-world structure and $W$ a set of literals. We first 
observe that the G-reduct $P^{\la \cA,W\ra}$ can be described as the
result of a certain two-step process. Namely, we define the 
\emph{epistemic reduct} of $P$ with respect to $\cA$ to be the disjunctive
logic program $P^\cA$ obtained from $P$ by removing every rule whose
epistemic premise $E$ satisfies $\cA\not\models E$, 
and by removing the epistemic
premise from every other rule in $P$. This construction is the three-valued
counterpart to the one we employ in our approach.
It is clear that the epistemic reduct of $P$ with respect to $\cA$, with 
some abuse of notation we will denote it by $\gl{P}{\cA}$, is a disjunctive
logic program with strong negation. 

Let $Q$ be a disjunctive program with strong negation and $W$ a set 
of literals. By the 
\emph{supp-reduct} of $Q$ with respect to $W$, $R^{sp}(Q,W)$, we mean 
the set of the heads of all rules whose bodies are satisfied by $W$ 
(which in the three-valued setting means that every literal in the body 
not in the scope of $\n$ is in $W$, and every literal in the body in 
the scope of $\n$ is not in $W$). A consistent set $W$ of literals is 
a supported answer set of $Q$ if $W\in\Min(R^{sp}(Q,W))$ (this is a 
natural extension of the definition of a supported model \cite{abw87,bg93}
to the case of disjunctive logic programs with strong negation; again,
we do not regard inconsistent sets of literals as supported answer sets).

Clearly, $P^{\la \cA,W\ra} = R^{sp}(\gl{P}{\cA},W)$. Thus, $\cA$ is a
world view of $P$ according to the definition by Gelfond if and only if
$\cA$ is a collection of all supported answer sets of $\gl{P}{\cA}$.  

We also note that by Proposition \ref{prop:2}, if $E$ is an epistemic
premise, then $\cA\models E$ if and only if $\cA'\models \sigma(E)$.
It follows that $\sigma(\gl{P}{\cA}) = \gl{\sigma(P)}{\cA'}$. In other 
words, constructing
the epistemic reduct of $P$ with respect to $\cA$ and then translating  
the resulting disjunctive logic program with strong negation into the
corresponding disjunctive logic program without strong negation yields 
the same result as first translating the epistemic program (in the
Gelfond's system) into our language of epistemic programs and then computing
the reduct with respect to $\cA'$. We note that there is a one-to-one
correspondence between supported answer sets of $\gl{P}{\cA}$ and supported
models of $\sigma(\gl{P}{\cA})$ ($\sigma$, when restricted to programs
consisting of rules without epistemic premises, is the standard transformation
eliminating strong negation and preserving the stable and supported 
semantics). Consequently, there is a one-to-one
correspondence between supported answer sets of $\gl{P}{\cA}$ and supported
models of $ \gl{\sigma(P)}{\cA'}$ (cf. our observation above). Thus, 
$\cA$ consists of supported answer 
sets of $\gl{P}{\cA}$ if and only if $\cA'$ consists of supported models
of $\gl{\sigma(P)}{\cA'}$. Consequently, $\cA$ is a world view of $P$
if and only if $\cA'$ is an epistemic supported model of $\sigma(P)$.
\hfill$\Box$ 

\section{Epistemic Models of Arbitrary Theories}

So far, we defined the notions of epistemic models, epistemic stable
models and epistemic supported models only for the case of epistemic 
programs. However, this restriction is not essential. We recall that 
the definition of these three epistemic semantics consists of two 
steps. The first step produces the reduct of an epistemic program $P$
with respect to a possible-world structure, say $\cA$. This reduct 
happens to be (modulo a trivial syntactic transformation) a standard 
disjunctive logic program in the language $\cL$ (no modal atoms 
anymore). If the set of models (respectively, stable models, supported 
models) of the reduct program coincides with $\cA$, $\cA$ is an 
epistemic model (respectively, epistemic stable or supported model) 
of $P$. However, the concepts of a model, stable model and supported 
model are defined for \emph{arbitrary} theories in $\cL$. This is 
obviously well known for the semantics of models. The stable-model 
semantics was extended to the full language $\cL$ by Ferraris 
\cite{fer05} and the supported-model semantics by Truszczynski
\cite{tr10}. Thus, there is no reason precluding the extension of the
definition of the corresponding epistemic types of models to the general
case. We start be generalizing the concept of the reduct.

\begin{definition}
Let $T$ be an arbitrary theory in $\cL_K$ and let $\cA$ be a
possible-world structure. The \emph{epistemic reduct} of $T$ with
respect to $\cA$, $\gl{T}{\cA}$ in symbols, is the theory obtained
from $T$ by replacing each maximal modal atom $K\vph$ with $\top$,
if $\cA\models K\vph$, and with $\bot$, otherwise.
\end{definition}

We note that if $T$ is an epistemic program, this notion of the reduct
does not coincide with the one we discussed before. Indeed, now no rule 
is dropped and no modal literals are dropped; rather modal atoms are 
replaced with $\top$ and $\bot$. However, the replacements are executed
in such a way as to ensure the same behavior. Specifically, one can show
that models, stable models and supported models of the two reducts 
coincide.

Next, we generalize the concepts of the three types of epistemic models.

\begin{definition}
\label{def14}
Let $T$ be an arbitrary theory in $\cL_K$. A possible-world
structure $\cA$ is an \emph{epistemic model} (respectively, an
\emph{epistemic stable model}, or an \emph{epistemic supported model})
of $P$, if $\cA$ is the set of models (respectively, stable models or
supported models) of $\cM(\gl{P}{\cA})$. 
\end{definition}

From the comments we made above, it follows that if $T$ is an epistemic 
program, this more general definition yields the came notions of epistemic
models of the three types as the earlier one. 

We note that even in the more general setting the complexity of reasoning
with epistemic (stable, supported) models remains unchanged. Specifically,
we have the following result.

\begin{theorem}
\label{thm:4}
The problem to decide whether an epistemic theory $T\subseteq\cL_K$
has an epistemic stable model is $\Sigma_3^P$-complete. The problem to 
decide whether an epistemic theory $T\subseteq\cL_K$ has an epistemic 
model (epistemic supported model, respectively) is $\Sigma_2^P$-complete.
\end{theorem}
Proof(Sketch): The hardness part follows from our earlier results concerning
epistemic programs. To prove membership, we modify Proposition
\ref{prop:char}, and show a polynomial time algorithm with a 
$\Sigma_2^P$ oracle (NP oracle for the last two problems) that
decides, given a propositional theory $S$ and a modal formula $K\vph$
(with $\vph\in\cL_K$ and not necessarily in $\cL$) whether $\cSM(S)
\models K\vph$ (respectively,  $\cM(S)\models K\vph$, or  $\cSP(S)
\models K\vph$). \hfill$\Box$

\section{Discussion}

In this paper, we proposed a two-valued formalism of epistemic 
theories --- subsets of the language of modal propositional logic.
We proposed a uniform way, in which semantics of propositional 
theories (the classical one as well as nonmonotonic ones: stable 
and supported) can be extended to the case of epistemic theories.
We showed that the semantics of epistemic supported models is closely
related to the original semantics of epistemic specifications proposed
by Gelfond. Specifically we showed that the original formalism of Gelfond
can be expressed in a straightforward way by means of epistemic programs
in our sense under the semantics of epistemic supported models. Essentially
all that is needed is to use fresh symbols $x'$ to represent strong 
negation $\neg x$, and use the negation operator of our formalism, 
$\vph \rra \bot$ or, in the shorthand, $\neg \vph$, to model the default 
negation $\n \vph$.   

We considered in more detail the three semantics mentioned above. 
However, other semantics may also yield interesting epistemic
counterparts. In particular, it is clear that Definition \ref{def14}
can be used also with the minimal model semantics or with the 
Faber-Leone-Pfeifer semantics \cite{flp04}. Each semantics gives rise
to an interesting epistemic formalism that warrants further studies.

In logic programming, eliminating strong negation does not result in 
any loss of the expressive power but, at least for the semantics of
stable models, disjunctions cannot be compiled away in any concise way 
(unless the polynomial hierarchy collapses). In the setting of 
epistemic programs, the situation is similar. The strong negation can be 
compiled away. But the availability of disjunctions in the heads and 
the availability of epistemic premises in the bodies of rules are 
essential. Each of these factors separately brings the complexity one 
level up. Moreover, when used together under the semantics of epistemic 
stable models they bring the complexity two levels up. This points to 
the intrinsic importance of having in a knowledge representation 
language means to represent indefiniteness in terms of disjunctions, 
and what is known to a program (theory) --- in terms of a modal operator 
$K$.

%%% See after end{doc}

\section*{Acknowledgments}
\vspace*{-0.1in}
This work was partially supported by the NSF grant IIS-0913459.

{\small
%\bibliographystyle{splncs}
%\bibliography{/home/mirek/Desktop/main.dir/docs.dir/nonmonlog}

\begin{thebibliography}{10}

\bibitem{gl90b}
Gelfond, M., Lifschitz, V.:
\newblock Classical negation in logic programs and disjunctive databases.
\newblock New Generation Computing \textbf{9} (1991)  365--385

\bibitem{gl88}
Gelfond, M., Lifschitz, V.:
\newblock The stable semantics for logic programs.
\newblock In: Proceedings of the 5th {I}nternational {C}onference on {L}ogic
  {P}rogramming (ICLP 1988), MIT Press (1988)  1070--1080

\bibitem{Gelfond91}
Gelfond, M.:
\newblock Strong introspection.
\newblock In: Proceedings of AAAI 1991. (1991)  386--391

\bibitem{FaberW09}
Faber, W., Woltran, S.:
\newblock Manifold answer-set programs for meta-reasoning.
\newblock In Erdem, E., Lin, F., Schaub, T., eds.: Logic Programming and
  Nonmonotonic Reasoning, 10th International Conference, LPNMR 2009. Volume
  5753 of Lecture Notes in Computer Science., Springer (2009)  115--128

\bibitem{mt88}
Marek, W., Truszczy\'{n}ski, M.:
\newblock Autoepistemic logic.
\newblock Journal of the {ACM} \textbf{38} (1991)  588--619

\bibitem{eg95}
Eiter, T., Gottlob, G.:
\newblock On the computational cost of disjunctive logic programming:
  propositional case.
\newblock Annals of Mathematics and Artificial Intelligence \textbf{15} (1995)
  289--323

\bibitem{abw87}
Apt, K., Blair, H., Walker, A.:
\newblock Towards a theory of declarative knowledge.
\newblock In Minker, J., ed.: Foundations of deductive databases and logic
  programming, Morgan Kaufmann (1988)  89--142

\bibitem{bg93}
Baral, C., Gelfond, M.:
\newblock Logic programming and knowledge representation.
\newblock Journal of Logic Programming \textbf{19/20} (1994)  73--148

\bibitem{bradix96jlp1}
Brass, S., Dix, J.:
\newblock {C}haracterizations of the {D}isjunctive {S}table {S}emantics by
  {P}artial {E}valuation.
\newblock Journal of Logic Programming \textbf{32(3)} (1997)  207--228

\bibitem{Inoue98-JLP}
Inoue, K., Sakama, C.:
\newblock Negation as failure in the head.
\newblock Journal of Logic Programming \textbf{35} (1998)  39--78

\bibitem{fer05}
Ferraris, P.:
\newblock Answer sets for propositional theories.
\newblock In: Logic Programming and Nonmonotonic Reasoning, 8th International
  Conference, LPNMR 2005. Volume 3662 of LNAI., Springer (2005)  119--131

\bibitem{tr10}
Truszczynski, M.:
\newblock Reducts of propositional theories, satisfiability relations, and
  generalizations of semantics of logic programs.
\newblock Artificial Intelligence (2010) In press, available through Science
  Direct at {http://dx.doi.org/10.1016/j.artint.2010.08.004}.

\bibitem{flp04}
Faber, W., Leone, N., Pfeifer, G.:
\newblock Recursive aggregates in disjunctive logic programs: semantics and
  complexity.
\newblock In: {Proceedings of the 9th European Conference on Artificial
  Intelligence (JELIA 2004)}. Volume 3229 of LNAI., Springer (2004)  200 -- 212

\end{thebibliography}

}
\end{document}